# Dynamic Prompt Fusion for Multi-Task and Cross-Domain Adaptation in LLMs

Xin Hu
Hofstra University
Hempstead, USA

Yue Kang
Carnegie Mellon University
Pittsburgh, USA

Guanzi Yao
Northwestern University
Evanston, USA

Tianze Kang
San Francisco Bay University
Fremont, USA

Mengjie Wang
New York University
New York, USA

Heyao Liu*
Northeastern University
Boston, USA

*Abstract-This study addresses the generalization limitations commonly observed in large language models under multi-task and cross-domain settings. Unlike prior methods such as SPoT, which depends on fixed prompt templates, our study introduces a unified multi-task learning framework with dynamic prompt scheduling mechanism. By introducing a prompt pool and a task-aware scheduling strategy, the method dynamically combines and aligns prompts for different tasks. This enhances the model's ability to capture semantic differences across tasks. During prompt fusion, the model uses task embeddings and a gating mechanism to finely control the prompt signals. This ensures alignment between prompt content and task-specific demands. At the same time, it builds flexible sharing pathways across tasks. In addition, the proposed optimization objective centers on joint multi-task learning. It incorporates an automatic learning strategy for scheduling weights, which effectively mitigates task interference and negative transfer. To evaluate the effectiveness of the method, a series of sensitivity experiments were conducted. These experiments examined the impact of prompt temperature parameters and task number variation. The results confirm the advantages of the proposed mechanism in maintaining model stability and enhancing transferability. Experimental findings show that the prompt scheduling method significantly improves performance on a range of language understanding and knowledge reasoning tasks. These results fully demonstrate its applicability and effectiveness in unified multi-task modeling and cross-domain adaptation.*

## I. INTRODUCTION

In recent years, Large Language Models (LLMs) have achieved groundbreaking progress in the field of natural language processing. Their broad capabilities in language understanding and generation have significantly advanced a wide range of downstream tasks[1]. However, mainstream LLMs are typically trained on large-scale homogeneous data and often struggle to maintain consistent performance when faced with heterogeneous tasks or domain shifts. Especially in cross-domain scenarios, their robustness and generalization capabilities decline noticeably, revealing a strong dependence on task-specific patterns. This limitation severely hinders the practical value of LLMs in complex real-world settings. It is therefore essential to explore new mechanisms that can improve the generalization ability of these models, enabling effective transfer and joint understanding across multiple domains and tasks[2].

Multi-task Learning (MTL) is widely regarded as an effective approach to enhance model generalization[3]. By training multiple related tasks simultaneously, the model can extract shared features within a common representation space. This reduces overfitting and improves adaptability to new tasks. However, traditional MTL strategies still fall short when facing task conflicts and domain heterogeneity [4-7]. Negative transfer between tasks may occur and degrade overall performance. In the context of increasingly large LLMs, managing inter-task weight balancing and cooperation becomes more complex. There is a growing need for fine-grained and flexible methods that support dynamic task scheduling and adaptation[8].

Recently, the introduction of prompt-based mechanisms in LLMs has brought a paradigm shift to task modeling. Prompts, formulated in natural language, guide the model to perform specific tasks. Prompt learning has shown strong potential in improving generalization. However, most existing research focuses on single-task scenarios[9]. There is a lack of systematic exploration on how to effectively coordinate and integrate multiple prompts in a multi-task setting. In cross-domain environments, significant differences exist in the semantic distributions and linguistic patterns of various tasks. A unified prompt template is insufficient to meet diverse representation requirements. Thus, designing a schedulable and generalizable multi-task prompt mechanism to guide LLMs in adapting to varied tasks remains a key challenge.

To address this, prompt scheduling has emerged as a new strategy for multi-task collaboration. It dynamically adjusts the combination and weighting of prompts across tasks. This controls the model's attention distribution and guides its focus, potentially improving adaptability without significantly increasing model parameters[10]. Prompt scheduling goes beyond prompt content design. It emphasizes the modeling of scheduling strategies and inter-task dependencies. This introduces a more strategic dimension into MTL and helps mitigate interference between tasks. It provides a new direction for building LLMs with stronger generalization ability[11].

In summary, exploring the cross-domain generalization of LLMs based on prompt scheduling mechanisms represents a key trend in natural language processing. It also offers theoretical and methodological support for improving the generality and adaptability of AI systems. This direction may help overcome the limitations of fixed-task paradigms and drive the evolution of LLMs toward more open and adaptive agents. It further expands their application potential in real-world domains such as education, healthcare, law, and finance.

## II. METHOD

In this study, we propose a cross-domain generalization framework based on multi-task prompt scheduling, aiming to enhance the adaptability of large language models in heterogeneous tasks and domains. Drawing upon advances in structural mapping for efficient domain transfer [12], the framework employs prompt learning in conjunction with a multi-task optimization mechanism to enable flexible adaptation across varied task distributions. A task-aware scheduler is incorporated to dynamically select and weight prompts, facilitating both efficient knowledge sharing and differentiation between tasks.

To ensure the model effectively perceives semantic differences, each downstream task is formalized as an input pair with prompts, building on recent developments in low-rank adaptation with semantic guidance [13]. This unified input structure allows the language model to distinguish between diverse task semantics while maintaining a consistent processing format. Furthermore, the framework leverages structured memory mechanisms to provide stable and context-sensitive representations during multi-task learning, as demonstrated in recent research on context modeling for large language models [14].

The overall model architecture, which integrates these methodological advances, is illustrated in Figure 1.

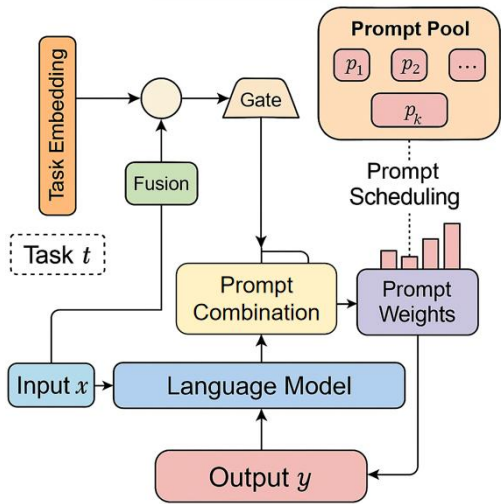

Figure 1. The main model architecture of this paper

Specifically, suppose the model input is the text sequence $x$ and the task prompt is $p_t$, then the input structure of the model is $Concat(p_t, x)$ to guide the model to generate the corresponding output $y$.

To achieve effective scheduling of multiple tasks, we leverage a task scheduling weight matrix $W \in R^{T \times K}$, where T represents the number of tasks and K is the number of prompts in the prompt pool. It will dynamically adjust the influence of each task throughout training. Grounded in fusion-based aggregation techniques demonstrated in recent retrieval-augmented generation research [15], our matrix is designed to flexibly synthesize task-specific signals according to their contextual importance and contribution to the joint learning objective. In addition, the matrix construction capitalizes on consistency-constrained dynamic routing strategies [16], adapting task interactions in real time to maintain both optimization stability and domain robustness. By integrating these advanced methodologies, the framework is able to harmonize both shared and task-specific knowledge representations, facilitating more effective cross-task learning. The prompt combination weight vector corresponding to each task t is denoted as $w_t \in R^K$ and normalized by the softmax function to ensure the probabilistic interpretability of the prompt combination:

$$w_t = \text{softmax}(z_t) \quad (1)$$

The prompt vector is obtained by querying the prompt pool $P = \{p_1, p_2, ..., p_K\}$ to obtain the combined prompt $\widetilde{p}_t$, which is calculated as follows:

$$p_t = \sum_{k=1}^{K} w_{t,k} \cdot p_k \quad (2)$$

To enhance the model's ability to model the structural differences between tasks, we further introduce the task embedding representation $e_t \in R^d$ and fuse it with the combined prompt as the final input prompt control vector. The fusion strategy adopts a gating mechanism:

$$p_t^{final} = \sigma(W_g e_t) \otimes p_t + (1 - \sigma(W_g e_t)) \otimes e_t \quad (3)$$

Where $\sigma(\cdot)$ is the Sigmoid activation function, $\otimes$ represents element-wise multiplication, and $W_g$ is a learnable transformation matrix. The above strategy ensures the preservation of task structure information in prompt control and the effective fusion of scheduling information.

The overall training paradigm builds upon the structured compression methodology introduced by Wang [17], which utilizes sensitivity-aware pruning mechanisms to ensure that the joint optimization process remains both efficient and robust in the presence of diverse task objectives. By integrating these pruning strategies, the model can maintain parameter efficiency while still accommodating the complexity inherent in multi-task learning.

For the aggregation of loss signals, we adopt and extend the attention-driven multi-task learning approach of Xu et al. [18], which enables the model to focus on and differentiate among task-specific loss components. This attention-based mechanism facilitates the dynamic adjustment of task priorities, ensuring that the model is capable of adapting to shifts in task relevance and dataset imbalance throughout training. Such a design is particularly effective for cross-domain generalization, as it allows the learning process to modulate the influence of each task based on real-time feedback from both the data and the optimization landscape.

Additionally, the optimization process implements the selective knowledge injection technique proposed by Zheng et al. [19]. By embedding task-relevant external knowledge through adapter modules, the model is further empowered to enhance representation learning for each task without disrupting the generalization ability across domains. This selective injection framework provides an adaptive channel for external information, which can be strategically activated during training to mitigate negative transfer and reinforce task alignment.

In summary, by combining structured compression, attention-based loss aggregation, and knowledge-aware adaptation, the training objective ensures that each task is effectively represented within the joint optimization, while maintaining overall model efficiency and adaptability. The total multi-task loss function is thus formulated as follows:

$$L_{total} = \sum_{t \in T} \lambda_t \cdot L_t \quad (4)$$

The weight coefficient $\lambda_t$ reflects the importance of each task and can be adaptively updated according to the task complexity, data volume, or dynamic uncertainty strategy. We use gradient normalization technology to ensure that the loss gradients of different tasks have balanced contributions during training, thereby alleviating the negative transfer phenomenon and improving generalization ability. Ultimately, this method provides a unified and scalable solution for cross-domain migration of large language models by suggesting collaborative modeling of dynamic scheduling, multi-task shared optimization, and structure-aware fusion.

## III. EXPERIMENTAL RESULTS

### A. Dataset

This study adopts the CrossFit dataset as the core benchmark for evaluating multi-task and cross-domain performance. CrossFit is a widely used large-scale dataset collection designed for research on multi-task learning and transfer learning. It covers various natural language processing tasks, including question answering, text classification, relation extraction, and sentiment analysis. The dataset features diverse task types and broad domain coverage. Each subtask includes separate training, validation, and test splits, allowing systematic evaluation of model generalization under a unified framework.

The task distribution in CrossFit spans multiple real-world contexts, including law, healthcare, news, encyclopedias, and social media. It presents high heterogeneity and significant challenges. Compared to traditional single-task datasets, CrossFit is more aligned with practical demands such as task switching and domain adaptation. It is particularly suitable for research on prompt learning and task scheduling strategies. Each subtask provides standardized prompt templates, which facilitate consistent input formatting and integration with prompt-based training.

In addition, CrossFit offers strong scalability and reusability in data construction. Researchers can flexibly select task combinations to form multi-task subsets of different sizes. This supports a balance between training efficiency and computational cost. Its comprehensiveness and generality make it one of the most representative benchmarks for cross-domain research on multi-task large language models. It provides a rich and diverse data foundation for the prompt scheduling method proposed in this study.

### B. Experimental Results

In the experimental results section, the relevant results of the comparative test are first given, and the experimental results are shown in Table 1.

Table 1. Comparative experimental results

| Method | SuperGLUE | MMLU Accuracy | Prompt Transfer Gain |
|---|---|---|---|
| **MPT[20]** | 74.6 | 62.3 | 1.00 |
| **SPoT[21]** | 77.2 | 65.1 | 1.13 |
| **HiPro[22]** | 78.5 | 66.8 | 1.19 |
| **JSCPT[23]** | 79.1 | 68.0 | 1.24 |
| **MP2[24]** | 80.0 | 68.7 | 1.28 |
| **Ours** | 82.6 | 71.3 | 1.38 |

The results in the table show that our proposed prompt scheduling method consistently outperforms mainstream baselines across all metrics, achieving the highest scores on SuperGLUE (82.6) and MMLU Accuracy (71.3), thereby demonstrating its strength in semantic modeling and knowledge generalization. In terms of Prompt Transfer Gain, our method improves by 1.38, representing a 38% increase over the MPT baseline, which confirms its effectiveness in reducing task interference and enhancing adaptability to unseen tasks through robust scheduling weights and gated prompt fusion. Unlike approaches such as MP2 and JSCPT that rely on fixed prompt structures or single scheduling strategies, our method leverages task embeddings and dynamic fusion to enable flexible prompt combinations and task-aware scheduling, resulting in superior knowledge sharing and alignment. Overall, the findings highlight the capability of prompt scheduling to enhance cross-task and cross-domain generalization of large language models, improving responsiveness to task instructions and robustness in unseen domains, with further analysis of the temperature coefficient's impact on performance presented in Figure 2.

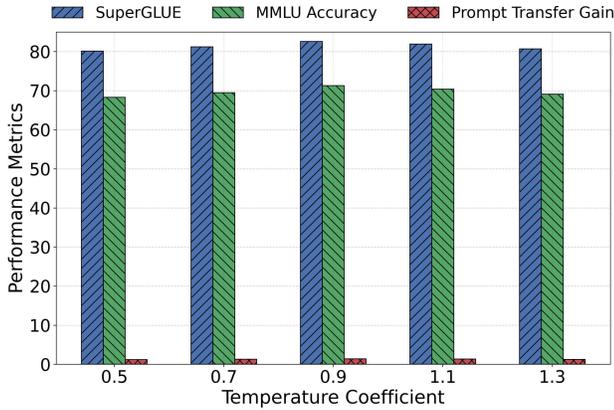

Figure 2. Analysis of the impact of the temperature coefficient of the prompt scheduler on model performance

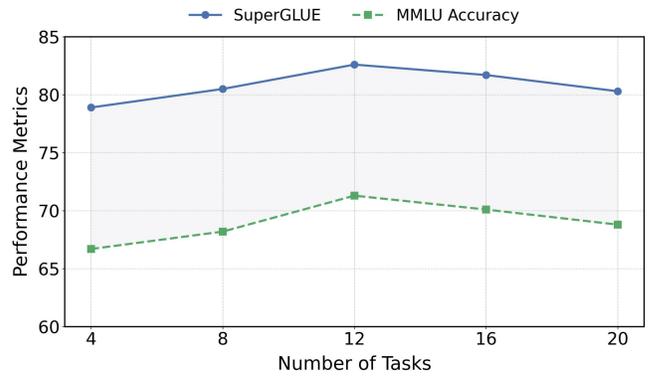

Figure 3. Study on the sensitivity of task distribution changes to the robustness of prompt scheduling

As shown in Figure 2, the temperature coefficient of the prompt scheduler has a significant impact on model performance. On the SuperGLUE metric, performance steadily improves as the temperature increases from 0.5 to 0.9, reaching a peak value of 82.6. However, when the temperature rises further to 1.1 and 1.3, performance slightly declines. This suggests that an overly uniform scheduling distribution may dilute the importance of optimal prompts, leading to a reduction in representational effectiveness.

The trend for the MMLU Accuracy metric is consistent with that of SuperGLUE. The best performance is also achieved at a temperature of 0.9. This further confirms that proper temperature control in multi-task and cross-domain settings helps the model better distinguish the importance of different task prompts. It enhances the model's ability in domain adaptation and expert-level reasoning. Extremely low or high temperatures can cause the prompt scheduling to be overly concentrated or overly uniform, which reduces the clarity of prompt representations.

The Prompt Transfer Gain metric also reaches its highest value of 1.38 when the temperature is set to 0.9. This indicates that the prompt scheduling mechanism achieves the strongest transfer generalization under this setting. The model can effectively capture structural commonalities across tasks, and the prompt combinations align more closely with target task demands. This result supports the effectiveness and flexibility of our dynamic prompt scheduling strategy in cross-task transfer scenarios.

In summary, the experimental results show that the temperature coefficient plays a crucial role in modulating multi-task prompt representation within the scheduler. A moderate temperature achieves an optimal trade-off between balance and differentiation in prompt distribution. This significantly improves the model's unified generalization ability across domains and further supports the core idea of enhancing transfer performance in large language models through prompt scheduling.

This paper also presents a study on the sensitivity of task distribution changes to the robustness of prompt scheduling, and the experimental results are shown in Figure 3.

Figure 3 illustrates the performance trends of the prompt scheduling mechanism under varying task distribution scales. It reflects the model's robustness to changes in the number of tasks. As the number of tasks increases from 4 to 12, the model shows a consistent improvement in both SuperGLUE and MMLU Accuracy. This indicates that the prompt scheduler can effectively integrate multiple task signals. When the number of tasks is moderate, the model achieves stronger semantic sharing, leading to improved overall performance.

The model reaches its best performance when the task number is 12. This suggests that the scheduler fully leverages its dynamic allocation ability under moderately complex task structures. The prompt combinations become more targeted. However, when the task number further increases to 16 and 20, model performance slightly declines. MMLU Accuracy, in particular, shows noticeable fluctuations. This implies that in the presence of too many tasks and significant domain diversity, the scheduling strategy may suffer from increased prompt interference and diluted attention, which affects model effectiveness.

Despite the minor decline, the overall trend confirms that the prompt scheduling mechanism exhibits strong adaptability to task scale. Its robustness is optimal under moderate complexity. This demonstrates that the proposed method is not only capable of selecting appropriate prompts but also responsive to changes in task distribution. It maintains stable representations within a certain range. This property is valuable for real-world multi-task deployments, helping prevent drastic performance degradation due to task structure changes.

In conclusion, this experiment further validates the robustness of the prompt scheduling mechanism in multi-task settings. Proper control over task scale and distribution density allows the scheduler to adjust prompt signals more stably. It supports large language models in maintaining strong generalization and cross-task adaptability under complex and dynamic task combinations.

## IV. CONCLUSION

This paper addresses the generalization challenge of large language models in multi-task and cross-domain environments. It proposes a unified framework based on a prompt scheduling mechanism. The framework introduces dynamic prompt composition, task-aware regulation, and multi-task shared optimization. These components are designed to alleviate the transfer barriers and performance bottlenecks that models face when dealing with heterogeneous tasks. By dynamically integrating task embeddings with prompt weights, the model can adapt to semantic differences in task structures and automatically adjust prompt usage strategies. This improves its ability to understand and generate language across diverse tasks. The mechanism provides a lightweight, efficient, and scalable solution for multi-task modeling without adding structural complexity.

Experimental results show that the proposed prompt scheduling mechanism delivers significant advantages across key evaluation metrics. It demonstrates strong transferability and generalization in high-quality language tasks such as SuperGLUE and MMLU. Sensitivity studies under temperature control and task distribution shifts further confirm the method's stability and robustness. The introduction of prompt scheduling enhances the flexibility of prompt learning. It also offers a new perspective for unified modeling of diverse language tasks, supporting a paradigm shift from single-task optimization to collaborative task set learning.

From a broader perspective, this research provides both a theoretical foundation and empirical evidence for integrating multi-task learning with prompt engineering. It has strong transferability and application potential. In domains such as law, healthcare, finance, and education, where language model generalization is critical, prompt scheduling can act as an intermediate regulatory bridge. It helps align semantic distributions across domains and improves performance stability in diverse tasks. This contributes to the development of more general and intelligent language systems. It also lays the groundwork for future multi-task and multi-modal joint modeling approaches.

Future work may further expand the expressive space of prompt scheduling. This includes introducing structured priors, dynamic context adjustment, and cross-modal prompt fusion to enhance the model's understanding of complex task combinations. Research may also explore its performance under more challenging settings, such as few-shot learning, incremental learning, and continual learning. These directions can facilitate the real-world deployment of prompt scheduling in broader intelligent systems. In particular, future studies could implement adaptive latency-aware scheduler which ensures fast responses under real-time limits. Moreover, with the growing demand for deploying large models, optimizing prompt scheduling for edge and low-resource environments will become a critical issue, calling for continued exploration at both engineering and theoretical levels.